\documentclass[3p,times]{elsarticle}

\usepackage{graphics}
\usepackage{lineno}
\usepackage{amsmath,mathrsfs}
\usepackage{makecell, multirow, diagbox}
\usepackage{url}
\usepackage{subfigure}

\usepackage{CJKutf8}

\usepackage{caption}
 \usepackage{booktabs}
\usepackage{algorithm}
\usepackage{algpseudocode}

\newtheorem{definition}{Definition}

\newdefinition{remark}{Remark}
\newproof{proof}{Proof}

\usepackage[
CJKbookmarks=true,
bookmarksnumbered=true,
bookmarksopen=true,
colorlinks,
pdfborder=001,
linkcolor=blue,
anchorcolor=blue,
citecolor=blue,
urlcolor=blue,
]{hyperref}

\usepackage[english]{babel}
\addto\extrasenglish{%
}

\modulolinenumbers[5]

\begin{document}

\begin{frontmatter}

\title{On Semi-Supervised Multiple Representation Behavior Learning}


\author[ictiip,amssmadis]{Ruqian Lu\corref{mycorrespondingauthor}}
\ead{rqlu@math.ac.cn}

\author[ictiip,ucas]{Shengluan Hou}
\ead{houshengluan1989@163.com}


\cortext[mycorrespondingauthor]{Corresponding author}

\address[ictiip]{Institute of Computing Technology  $\&$ Key Lab of IIP, Chinese Academy of Sciences, Beijing 100190, China}
\address[amssmadis]{Academy of Mathematics and Systems Sciences $\&$ Key Lab of MADIS, Chinese Academy of Sciences, Beijing 100190, China}
\address[ucas]{University of Chinese Academy of Sciences, Beijing 100049, China}

\begin{abstract}
We propose a novel paradigm of semi-supervised learning (SSL) – the semi-supervised multiple representation behavior learning (SSMRBL). SSMRBL aims to tackle the difficulty of learning a grammar for natural language parsing where the data are natural language texts and the `labels' for marking data are parsing trees and/or grammar rule pieces. We call such `labels' as compound structured labels which require a hard work for training. SSMRBL is an incremental learning process that can learn more than one representation, which is an appropriate solution for dealing with the scarce of labeled training data in the age of big data and with the heavy workload of learning compound structured labels. We also present a typical example of SSMRBL, regarding behavior learning in form of a grammatical approach towards domain-based multiple text summarization (DBMTS). DBMTS works under the framework of rhetorical structure theory (RST). SSMRBL includes two representations: text embedding (for representing information contained in the texts) and grammar model (for representing parsing as a behavior). The first representation was learned as embedded digital vectors called impacts in a low dimensional space. The grammar model was learned in an iterative way. Then an automatic domain-oriented multi-text summarization approach was proposed based on the two representations discussed above. Experimental results on large-scale Chinese dataset SogouCA indicate that the proposed method brings a good performance even if only few labeled texts are used for training with respect to our defined automated metrics.
\end{abstract}


\end{frontmatter}


\section{Introduction}
Usually machine learning programmers are embarrassed by the scarce of labeled training data. Experts then said `Don't worry. There are lots of source data which are unlabeled but useful'.  With this idea people have invented semi-supervised learning (SSL), which detects some common features of labeled and unlabeled examples to help determine the model characteristics. There are many different types of SSL. To name a few, we have seen semi-supervised classification learning (SSCL) \cite{kipf2016semi}, semi-supervised representation learning (SSRL) \cite{banijamali2016semi}, semi-supervised reinforcement learning (SSRinfL) \cite{finn2016generalizing}, semi-supervised behavior learning (SSBL) \cite{cheng2016semi,dai2015semi}, etc.

There is another situation which leads to the invention and development of SSL. This is not because of the lack of a massive set of training data, but because of the difficulty or complexity of training a model, or more precisely, because of the difficulty or complexity of training each sample of a model. Examples include: learning how to play Go by analyzing a set of Go playing records, learning how to prove a mathematical theorem by reading through a set of mathematical theorems proves, learning how to command a campaign by reading a few historical records of military campaigns, and finally, as we will introduce in a later section of this paper, how to learn a grammar of natural languages (in a certain domain) by manually parsing a limited set of natural language texts. The common property of these examples is the enormous cost of training a single example such that usually one cannot tolerate this high cost. 

Generally speaking, an SSL algorithm of the above type usually presents a multi-lateral style. Take the military campaign as an example. It tells officers how to conduct a campaign. This concerns human behavior, so it is SSBL. Officers always learn from their war practices. They get positive or negative lessons from practice. Their experiences should be fostered through many practical combats, so it is SSRinfL. Officers have to write strategic and tactic plans for each campaign before it and summaries after it. The general lessons acquired from these materials will be written into textbooks for training latter officers, so it is SSRL.

Technically, an SSL algorithm of the above type is often more difficult and complicated to design and to deal with. Firstly, it concerns complicated problems which generally don't have fixed and unique solutions. On the contrary, the solutions are usually heuristic and probabilistic even if a massive data set has been used for training. Secondly, the massive data set to be trained is usually growing steadily (a big data stream). So an incremental approach is needed. Thirdly, it usually involves multi-factors (multi-views), such that a single criterion is not enough for deciding the resulting model. Fourthly, given that the problem is of SSRL type, the algorithm is usually not limited to find a single representation. In that case, although a first representation is learned, the result is not yet the wanted final result. Often is a second or even more representations are needed, which play different roles in the learning process. We call this paradigm as semi-supervised multiple representation behavior learning (SSMRBL).

In this paper, we will present an example of SSMRBL. It is a project about domain-based multiple text summarization (DBMTS), which we have been working on since last years. Given a set $DS$ of natural language texts from some application domain $D$. A limited subset $T$ is selected from $DS$ as training data. Each text of $T$ will be parsed manually to get a parsing tree based on domain knowledge. During this process the programmer's behavior will be collected by computer to form a grammar, where the tree structures form a basis for reduction rules and the contexts of parsing will be recorded for building preference rules for resolving shift-reduce and reduce-reduce conflicts. The parsing tree of each text together with the set of reduction and preference rules contributed by it are considered as `label' of this text. The whole set of these `label's forms a grammar, considered as representation of the training set. Thus the learning is semi-supervised. To cope with `unlabeled' data, we introduce another representation–a digital vector space for text embedding. It makes use of `closeness' concept of the digital space to select the `most representative' texts for initial labeling, to transfer labels to unlabeled data and to select contents for automatic summarization.

\begin{figure}[t]
\centering
\includegraphics[width=0.65\textwidth]{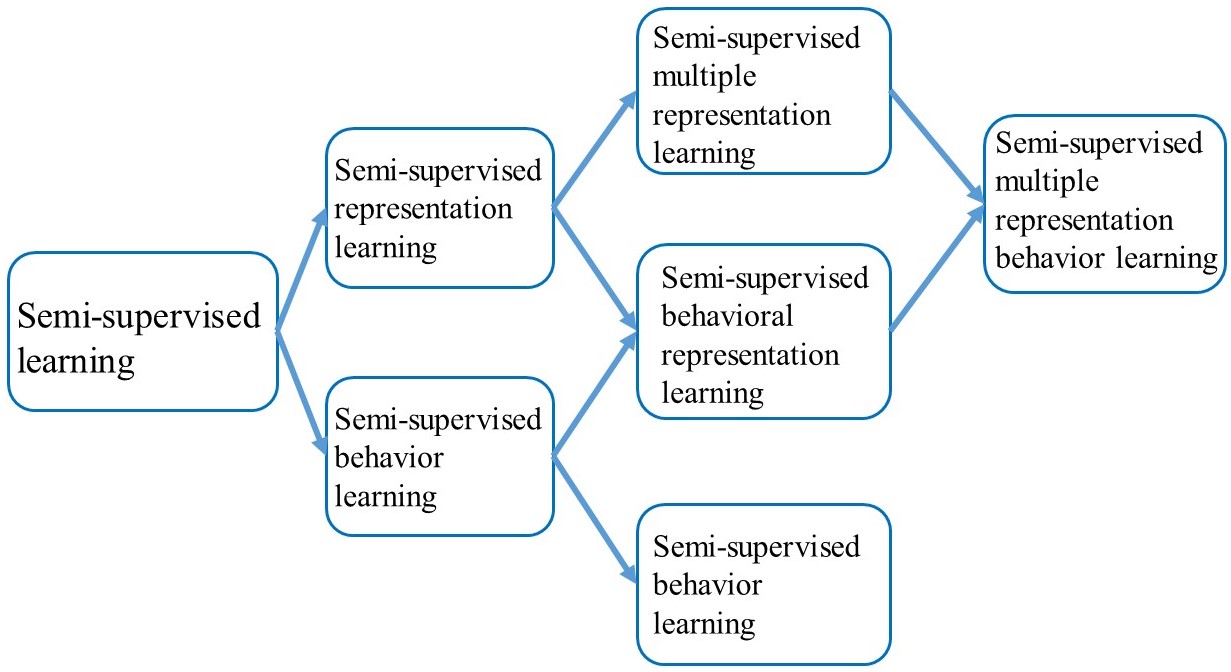}
\caption{The relations between different SSL schemas}
\label{fig_1_rel}
\end{figure}

In order to put forward the special characteristics of our approach, we also call it as semi-supervised behavioral representation learning, because learning a grammar is learning the way as how to analyze the texts of a particular domain. The relations between different SSL schemas are illustrated in \autoref{fig_1_rel}.

The contributions of this work are summarized as follows:
\begin{itemize}
\item We propose SSMRBL, a novel paradigm about incremental multiple representation learning that can learn more than one representation according to specific tasks. SSMRBL is an appropriate solution for dealing with the scarce of labeled data.
\item Under the framework of SSMRBL, we give an instance: a project about DBMTS. DBMTS here is a domain-oriented multi-text summarization architecture, which first parse texts into discourse trees and then extract summary units using a representative selection strategy.
\item We also incorporate rhetorical structure theory (RST) into DBMTS such that the learned grammar model is attributed rhetorical structure grammar (ARSG) and the basic processing unit is elementary discourse unit (EDU), which is a key concept in RST that often corresponds to a clause or a simple sentence.
\item To investigate the factors involved in representation learning and summary generation, we perform extensive experiments, propose novel automated metrics, and report findings on them.
\end{itemize}

The remainder of this paper is organized as follows. \autoref{sec_relatedworks} is related works. In \autoref{sec_textembedding}, learning of text embedding (i.e. the first representation) is detailed. \autoref{sec_arsg} and \autoref{sec_getinitalgrammar} are about attributed rhetorical structure grammar and how to obtain the initial grammar model. In \autoref{sec_secondrep}, we learn the grammar model in an iterative way. Domain-oriented multi-text summarization is introduced in \autoref{sec_mdts}. Experimental results and analysis are in \autoref{sec_experiments}. We conclude this paper in \autoref{sec_conclusion}.

\section{Related Works}
\label{sec_relatedworks}

Our research builds on previous works in the field of semi-supervised learning, representation learning, rhetorical structure theory, and multi-text summarization.

Semi-supervised learning aims to improve generalization on supervised tasks using unlabeled data \cite{chapelle2009semi,weston2012deep}. SSL usually combine unlabeled data with very smaller set of labeled data to gain better data representation or classification accuracy. There are wide range of applications, such as text classification \cite{ahmed2009sisc,kipf2016semi}, discourse relation classification \cite{Hernault2010A}, and object detection \cite{rosenberg2005semi}, etc. Based on both labeled and unlabeled instances, the problem of SSL is defined as learning a function from labeled data to unlabeled data. SSL contains two learning paradigms, transductive learning and inductive learning. Transductive learning directly apply the function on the unlabeled but observed instances at training time. On the other hand, the aim of inductive learning is to learn a parameterized function that is generalizable to unobserved instances. The basic assumption of SSL is that nearby nodes tend to have the same labels \cite{yang2016revisiting}.

Representation learning is the process of learning representations of the data that make it easier to extract useful information when building classifiers or other predictors. A word embedding, which was first introduced by Bengio et al. \cite{bengio2003neural}, is a continuous vector representation that captures semantic and syntactic information about a word. Word embeddings have proven useful in many NLP tasks. Embeddings for longer texts (phrase, sentence and even texts) are also necessary in many circumstances. For deriving them based on word embedding, there are many methods. For instance: (1) Vector addition: a simple way that consider it as the sum of all words without regarding word orders; (2) Recurrent neural network: a text embedding is the concatenation of the output states of a RNN or bidirectional RNN over its word embeddings, in which LSTM or GRU is commonly used \cite{kedzie2018content}; (3) Convolutional neural network: a series of convolutional feature maps are used over word embedding matrix, the text representation is the concatenation of all the convolutional filter outputs after max pooling over time \cite{kim2014convolutional}.

Note that rhetorical structure theory (RST) is a theory proposed by Mann and Thompson in last century for text structuring \cite{mann1988rhetorical,Taboada2006Rhetorical}. RST aims to investigate how clauses, sentences and even text spans connect together into a whole in a logical and topological way. RST explains text coherence by postulating a hierarchical, connected tree-structure (denote as RS-tree) for a given text, in which every part has a role, a function to play, with respect to other parts in the text. Each leaf node in an RS-tree is called an elementary discourse unit (EDU). RST assumes that each EDU (or text span) of a natural language sentence (or text) may either be a nucleus or a satellite. An RS-tree is a binary tree. Its two branches are either a pair (nucleus, satellite) or two nuclei. Rhetorically, the nucleus is always in the center and considered as more significant, while the satellite is used to modify the nucleus and as such is less significant. RST has been widely used in natural language analysis, understanding and also generation that need to combine meanings of larger text units.

According to the number of input texts, text summarization can be broadly categorized into single-text summarization (SDS) and multi-text summarization (MDS). MDS is more complicated since there are more problems should be tackled, such as contradiction, redundancy and complementarity, when generating summarization from multiple texts. For decades, there are various kinds of approaches have been proposed, including graph-based methods \cite{canhasi2014graph,mihalcea2004textrank}, lexical chain-based methods \cite{Barzilay1997,chen2005automatic,hou2017holographic}, constraint optimization-based methods \cite{berg2011jointly,gillick2009scalable,mcdonald2007study}, traditional machine learning-based methods \cite{fattah2014hybrid,gambhir2017recent}, and deep learning-based methods \cite{cheng2016neural,peyrard2019simple,rush2015neural}, etc. Deep learning-based approaches are able to leverage large-scale training data and have achieved competitive or better performance than traditional methods with the availability of large-scale corpora. The above methods, on the other hand, can also be grouped into abstractive ones and extractive counterparts. In contrast to extractive summarization where a summary is composed of a subset of sentences or words lifted from the input texts, abstractive summarization concerns the generation of new sentences, new phrases while retaining the same meaning as the same source text(s) have, which are more complex than extractive ones. Extractive approaches are now the mainstream ones.

This work is a novel framework that mainly based on the above related works, which attempts to teach the system to parse texts into discourse structure trees based on small set of human annotated texts and then to determine whether an EDU should be selected to produce the final summary of multiple texts. In this way, the summary generated by our method is the extractive one.

\section{Text Embedding: Learn the First Representation}
\label{sec_textembedding}

The first representation of domain texts is given as a vector set in a target digital space. Each text of the corpus is mapped to a vector in this space. Two texts are similar if their mappings in this space are close neighbors. Usually a domain related text introduces a review on the current state of the related domain. The major content of such texts is manifested by comments and reviews consisting of a set of domain concepts and expressions, which also form the basis of grammar reduction in our ARSG \cite{lu2019attributed}.

In this paper, a lexical core is defined as a quadruple of domain concepts for representing domain news or reviews. Its semantics can be summarized as (When, Who/Where, What, hoW) which are the well-known five ``W''s factors for news reports. Following is an example. Assume we have the following text: 

\textit{(News: 2019 January) Today, IMF noticed a downward risk of global economy. The growth has surpassed its peak value. IMF has reduced growth anticipation of developed, new markets and developing economies resp. A UNO's report pointed out that USA’s economy growth rate will be going down to 2.5\% in 2019 and 2.2\% in 2020, while EU’s growth rate will continue to be 2\% and China's rate from 6.9\% will go down to 6.3.}

This text contains the following lexical cores:

(2019, IMF notice, global economy, downward risk), (2019, IMF anticipation, growth of economies, reduced), (2020, USA, economy growth rate, going down), (2020, EU, economy growth rate, continue), (2020, China, economy growth rate, go down)

After careful analysis on many articles about news reports, we found that most of the articles have time information. Time information often present in the beginning of these articles, most of which have only one time information and a small number of them have one more time changes. These time changes imply the object changes of an event in most cases. \autoref{tab_excerptsnews} shows the excerpts of news articles. On the other hand, time information is also the important component about the domain news or reviews with respect to its semantics. It's useful and necessary to import time information into lexical cores. Based on the above observations, we empirically take time information as a component of the lexical core.

\begin{table}[!t]
\caption{Excerpts of news reports}
   \centering
   \begin{tabular}{p{150pt}p{260pt}}
   \toprule
   Chinese Text & English Translation \\
   \midrule
\begin{CJK}{UTF8}{gkai}
　　2019年9月1日，工业和信息化部副部长陈肇雄应邀作客第142期“中浦讲坛”，作“加快推动制造业数字化转型”专题报告。

　　陈肇雄指出，新中国成立70年来，我国制造业保持持续快速发展，为我国经济建设作出突出贡献。特别是党的十八大以来，我国制造业发展实现了历史性突破、取得了历史性成就，综合实力稳步提升、创新能力明显增强、产业结构加快调整、发展环境不断优化、开放水平大幅提升，为实现“两个一百年”目标奠定了坚实基础。

　　......
\end{CJK}
 & 
On September 1, 2019, Chen Xiongxiong, deputy minister of MIIT, was invited to be the guest of the 142th ``Zhongpu Forum" and gave a report entitled ``Accelerating the Digital Transformation of Manufacturing Industry".

Chen Shaoxiong pointed out that since the founding of China, our country's manufacturing industry has maintained sustained and rapid development and made outstanding contributions to China's economic construction. Especially since the 18th National Congress of the Communist Party of China, the manufacturing industry has achieved historic breakthroughs and achieved historic achievements. Its comprehensive strength has been steadily improved, its innovation capability has been significantly enhanced, its industrial structure has been accelerated, its development environment has been continuously optimized, and its openness has been greatly improved. Solid foundation has been laid for achieving the ``two hundred years" goal.

...... \\
\midrule
\begin{CJK}{UTF8}{gkai}
　　国家统计局天津调查总队最新监测数据显示：8月份我市CPI同比上涨2.4\%。其中，食品价格上涨4.8\%，非食品价格上涨1.9\%；消费品价格上涨2.3\%，服务价格上涨2.5\%。1—8月平均，我市CPI比上年同期上涨2.2\%。

　　8月份我市CPI环比上涨0.5\%。其中，食品价格上涨1.7\%，非食品价格上涨0.3\%；消费品价格上涨0.7\%，服务价格上涨0.3\%。
\end{CJK}
 & 
The latest monitoring data of the Tianjin Investigation Team of National Statistical Bureau shows that the CPI of the city in August increased by 2.4\% year-on-year. Among them, food prices rose by 4.8\%, non-food prices rose by 1.9\%; consumer prices rose by 2.3\%, and service prices rose by 2.5\%. From January to August, the city's average CPI increased by 2.2\% over the same period of the previous year.

In August, the city's CPI rose by 0.5\%. Among them, food prices rose by 1.7\%, non-food prices rose by 0.3\%; consumer prices rose by 0.7\%, and service prices rose by 0.3\%.
\\
   \bottomrule
\end{tabular}
\label{tab_excerptsnews}
\end{table}

The procedure of embedding is as follows: the computer scans each text to extract all its lexical cores. We use $q = (t, a, o, s)$ to denote any lexical core, where $t, a, o, s$ means time, agent, object and state change respectively. Since our attention is focused on domain oriented texts, the design of lexical cores has in particular taken those terms in consideration, which occur most frequently in texts of type domain news. For example, we will put more weight to the five ``W''s. Among the four components of a lexical core, $t, a, o, s$ corresponds to `when', `who' (or `where'), `what' and `how' resp. Each of these components is represented as a phrase, such as `price-of-rice'.

An unified function $g$ maps each lexical core to a vector in a $4\times w$ dimensional space $R$ with phrase embedding technique the requirement that $g ((t, a, o, s)) = g(q) = q' = <t', a', o', s'>$, where $q'$ is a vector in $R$, while $t', a', o', s'$ are its four components, each of which is a $w$-dimensional vector, where $w$ is the dimension size of phrase embedding. 

Given that each lexical core is represented by a vector in space $R$ and each text usually provides several lexical cores, the embedding of the whole text is represented by the addition, called impact, of all vectors representing its lexical cores. Although the lexical core and impact vectors are in the same space $R$, we consider them as being in two separate spaces for easing the discussion. We call the space for lexical cores (impacts) as core space (impact space) resp.

\begin{algorithm}[h] 
\caption{Embed texts in a vector space} 
\label{alg_1_embedtexts}
\begin{algorithmic}[1] 
\Require 
A set $D$ of natural language texts from an application domain.
\State Extract lexical cores in form of $(t, a, o, s)$ from EDUs of each text;
\State Embed each lexical core to a vector simply called core in a $4\times w$-dimensional space $R$ by embedding the components $t, a, o, s$ as the $4\times w$ coordinates of the lexical core, where $w$ is the dimension of phrase embedding;
\State For each text calculate the impact of all its core vectors. This impact is considered as the embedded mapping of the text.
\end{algorithmic} 
\end{algorithm}

\begin{figure}[t]
\centering
\includegraphics[width=0.55\textwidth]{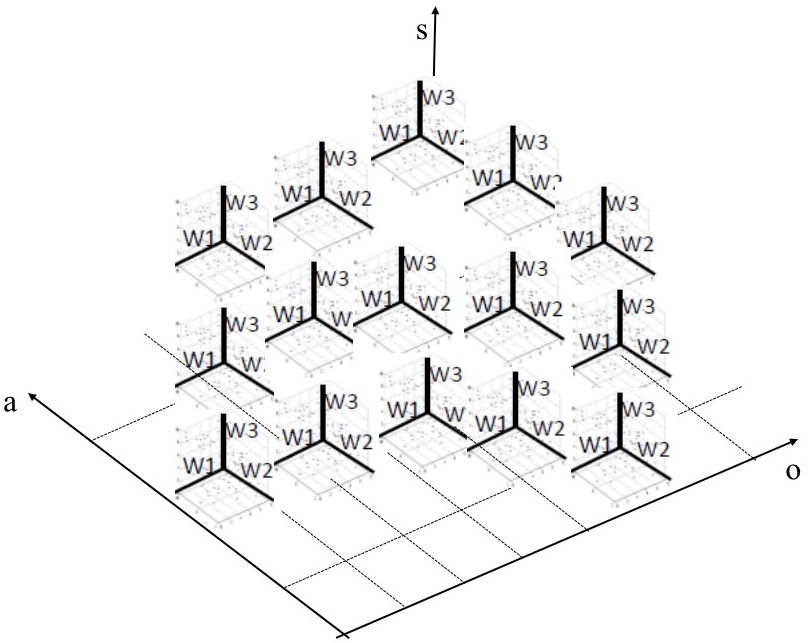}
\caption{The $4\times w$ dimensional space for cores and impacts as vectors}
\label{fig_2_space}
\end{figure}

\autoref{fig_2_space} is a simplified illustration of core space, where we assume $w=3$ and omit for the moment the first component t of a core. Note that if we only consider impacts of the cores, then the space is an impact space.

For the four components of the same lexical core, the equation
\begin{equation*}
t + a + o = s
\end{equation*}
is fulfilled. Thus the loss function for training lexical core embedding is
\begin{equation}
L=\sum\limits_{(t,a,o,s)\in W}\sum\limits_{(t',a',o',s')\in W'}[\gamma+d(t+a+o,s)-d(t'+a'+o',s')]_+
\tag{$100$}
\label{eq_emb_loss}
\end{equation}
\begin{equation}
W'=\bigcup\limits_{(t,a,o,s)\in W}\{(t',a,o,s),(t,a',o,s),(t,a,o',s),(t,a,o,s')\}
\tag{$101$}
\end{equation}
where $t',a',o',s'$ correspond to corrupted $t, a, o, s$ respectively.

\begin{figure}[t]
\centering
  \subfigure[Impacts of lexical cores of domain texts]{\includegraphics[width=2.2in]{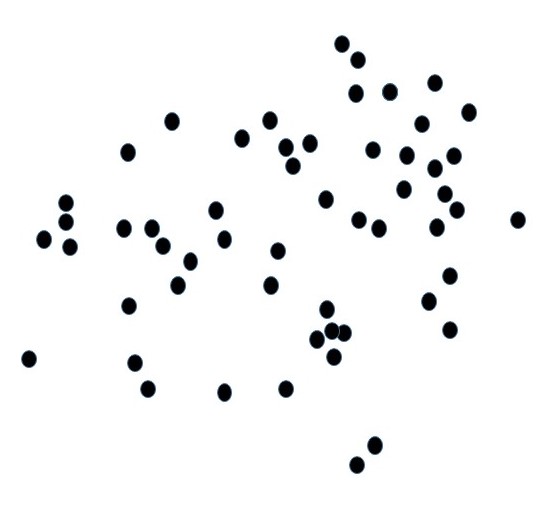}}
  \subfigure[Impacts distributed in cubes]{\includegraphics[width=2.3in]{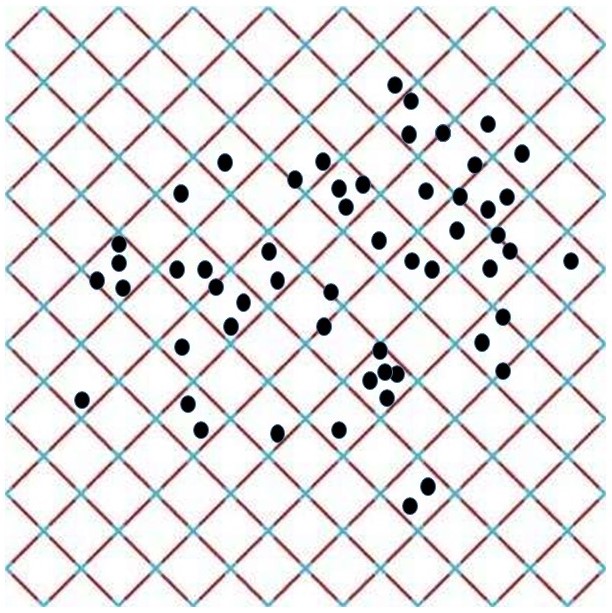}}
\caption{Text embedding-The first representation}
\label{fig_3_embedding}
\end{figure}

\autoref{fig_3_embedding} is an illustration of text embedding. Each text is represented by an impact vector in the vector space, whose coordinates are computed by the trained lexical cores according to \autoref{alg_1_embedtexts}.

In this way, the embedding for each domain text can be derived, which is the basis of the following procedures.

\section{Attributed Rhetorical Structure Grammar—A Behavioral Representation}
\label{sec_arsg}
In a previous paper \cite{lu2019attributed} we have introduced an approach of attribute grammar-based text summarization. The main idea is as follows. Assume a large set of texts is given relating to an application domain $D$. Build a domain knowledge base $DKB$ of domain concepts and domain relations. Consider them as terminal and non-terminal symbols of a grammar. Further use rhetorical relations as `glue' to build production rules from these terminal and non-terminal grammar symbols text by text. Then introduce different kinds of attributes, including the rhetorical relations, to make the grammar an attributed one. In this way we get an attributed rhetorical structure grammar.

We start from the definition of a simple version (without probability) of this attribute grammar.

\begin{definition}
A simple attributed rhetorical structure grammar (SARSG) is a six tuple: $(RS, DRE, KCP, RRE, PR (AT), AT)$, where $RS$ is the start symbol, $DRE$ the set of domain relations, $KCP$ the set of domain concepts, $RRE$ the set of rhetorical relations, $AT$ the synthesized attributes, where each attribute is an (arithmetical or logical) function with grammar symbols as arguments. $PR(AT)$ is the set of production rules attached with attribute equations, $\{RS\}\bigcup DRE$ the non-terminals, $RRE\bigcup KCP$ the terminals.
\label{def_sarsg}
\end{definition}

A production rule of $PR(AT)$ has the following form:
\begin{equation}
(ae): D(Lf)\leftarrow A(X),B(Y)
\label{eq_rule_attr}
\end{equation}
where $ae$ is the attribute equations, denoting how the values of $D$'s attributes are calculated from those of $A$ and $B$'s attributes. $D$ is the left side (parent) symbol, $A$ and $B$ the right side (child) symbols. $(X, Y)$ is $(N, S)$ or $(S, N)$ or $(N, N)$, where $N$ means nucleus and $S$ means satellite. $A, B, D$ are domain relations, $Lf$ the reason (a logic formula) of confirming this rule as a legal production rule reducing the string $AB$ to $D$ under the current parsing context. 

The attribute equation $ae$ is a function calculating attribute values of the production's left symbol $D$ with attribute values of same production's right side as arguments.

\begin{definition}
An attributed rhetorical structure grammar (ARSG) is a seven tuple: $(RS, DRE, KCP, RRE, PPR(AT), PF(AT), AT)$, where $RS, DRE, KCP, RRE$ and $AT$ are the same as in SARSG. $PPR(AT)$ is the set of probabilistic production rules, attached with attributes and reasons (see Definition \autoref{def_sarsg} and below), $PF(AT)$ the set of shift-reduce precedence rules. 
\end{definition}

A production rule of $PPR(AT)$ has the following form:
\begin{equation}
(ae): D(Lf)\leftarrow A(X),B(Y) \qquad n
\end{equation}
That means it is a production rule of SARSG followed by a positive integer $n$ for calculating the probability of reducing $AB$ to $D$ dynamically (in case reduction is required). More precisely, if there are $k$ productions 
$$(ae): D_i(Lf)\leftarrow A(X),B(Y) \qquad n_i, 1\leq i \leq k,$$
then the probability that the $j$-th production will be chosen is equal to $\frac{n_j}{\sum_{1\leq i \leq k} n_i}$.

$PF(AT)$ is a set of precedence tuples where each tuple is in the form
\begin{equation}
(A, B, C, \prec, slf_{A, B, C}, p_s) \quad or \quad (A, B, C, \succ, rlf_{A, B, C}, p_r)
\label{eq_prec_tup}
\end{equation}
where $ABC$ is a string of three neighboring grammar symbols during parsing. $slf_{A, B, C}$ (short for shift logic formula) is a reason for shifting the parser over $C$, while $rlf_{A, B, C}$ (short for reduce logic formula) is a reason for reducing $(A, B)$ to some $DRE$. The truth values of both $slf_{A, B, C}$ and $rlf_{A, B, C}$ depend on the attribute values of $A, B$ and $C$. $p_s$ and $p_r$ are probabilities with  $p_s+p_r=1$. Thus the six-tuples (\ref{eq_prec_tup}) are used for resolving shift-reduce conflicts.

\section{Training a Sample Set to get the Initial Grammar Model}
\label{sec_getinitalgrammar}

The basic idea is as follows: We first decide a target application domain $D$. With the aid of some domain oriented dictionaries and thesaurus we build a knowledge base $KB$ of $D$, including domain concepts and domain relations, where the domain concepts are abstract or concrete entities, while domain relations are states or state changes of domain concepts. For example, if the domain is world economy and trade (WET), then its domain concepts may be market, price, stock, country, bank etc. while inflation, balance, improved, expended, etc. are its domain relations. Given a set $DS$ of texts in $D$, we want to learn a grammar $G$ (see \autoref{sec_arsg}) from $DS$ based on $KB$, such that other texts of $D$ may be parsed by $G$ to produce a parsing tree for each text. All domain concepts (relations) are considered as terminal (non-terminal) symbols of $G$, which is context sensitive and designed as an attributed grammar, where the rhetorical relations are the most important attributes. All grammar rules are binary with a rhetorical relation for parent node and attributes $[nucleus, satellite]$ or $[nucleus, nucleus]$ for the two child nodes. Accordingly, the parsing result is a binary tree (parsing tree) called an attributed rhetorical structure tree.

Given a subset $DS1$ of $DS$ for training, the process of parsing texts in $DS1$ is done by a cooperation of programmer and computer. This process consists of two sub-processes. In the first sub-process, the programmer mimics a machine compiler by scanning the text from left to right. He/she decides a shift or a reduce whatever the right action is due, which transforms the text as a sentential step by step bottom up. The parsing tree is then constructed level by level. At the same time the computer observes, collects and records all information during every action done by the human programmer. In particular, if the computer detects that the human shows different behaviors within the same parsing context, it will produce a separate grammar rule for each behavior and record the number of its occurrences as a part of the rule. Finally, in the second sub-process, for each text of the training set there will be generated a parsing tree and a set of grammar rules. The computer synthesizes all these rule sets to form the wanted grammar. There may possibly be conflicting rules, which make the grammar a probabilistic one. The number of instances of each rule represents its weight. These numbers will be used during application of the grammar to calculate the probability for each rule to apply.

More precisely, in the first sub-process, each time when the programmer makes a decision of shift or reduce, the computer records this decision and the context of the sentential where the decision is made. If the decision is a reduce, the machine produces a rough reduction rule as shown in (\ref{eq_sim_rule})
\begin{equation}
D(Lf)\leftarrow A, B
\label{eq_sim_rule}
\end{equation}
where $Lf$ is the recorded context of the sentential during reduction. We call it the reason of reduction. At this time, the human parser may or may not assign additional information to (\ref{eq_sim_rule}) to enrich it into the form of (\ref{eq_rule_attr}). Note that if this additional information is missing, the default meaning can be represented as the following
\begin{equation}
(\text{straight forward}): D(Lf)\leftarrow A(\text{don't care}),B(\text{don't care})
\end{equation}
where ``strait forward'' means $D$ accepts all attribute values of $A$ and $B$, ``don't care'' means the rhetorical relations of $A$ and $B$ may take any valid values.

Note that no matter the decision is shift or reduce, a rough preference rule is always produced by the computer. It has the following form
\begin{equation}
(A, B, C, \prec, slf_{A, B, C}, ) \quad or \quad (A, B, C, \succ, rlf_{A, B, C}, )
\label{eq_prec_without_p}
\end{equation}
where $slf_{A, B, C}(rlf_{A, B, C})$ denotes the context when shift from $A, B$ to $A, B, C$ (reduction of $A$ and $B$ in front of $C$) is made. The blank spaces in parentheses are saved for possible probability values.

In the second sub-process, all instances of the same rule in form of (\ref{eq_rule_attr}) or (\ref{eq_prec_without_p}) are collected to calculate the weight. Regarding production rules in form of (\ref{eq_rule_attr}), if the parent $D$ of $A$ and $B$ is unique, then rule (\ref{eq_rule_attr}) gets the form
\begin{equation}
(ae): D(Lf)\leftarrow A(X),B(Y) \qquad 1
\end{equation}
It means the string $AB$ always reduces to $D$ if the decision is reduction. Otherwise, if there are several parents $D_1, D_2, \cdots, D_k$ for reducing $AB$, then we divide all such rules in $k$ groups such that all rules in the $i$-th group, $i = 1, 2, \cdots, k$, share the same parent symbol $D_i$. Assume the number of rule instances in the $i$-th group is $N_i$, then we obtain $k$ probabilistic rules
\begin{equation}
\begin{split}
(ae_1): D_1(Lf_1)\leftarrow A(X_1),B(Y_1) \qquad N_1 \\
(ae_2): D_2(Lf_2)\leftarrow A(X_2),B(Y_2) \qquad N_2 \\
\cdots \\
(ae_k): D_k(Lf_k)\leftarrow A(X_k),B(Y_k) \qquad N_k \\
\end{split}
\end{equation}
This means if a reduction of $AB$ is to be made during parsing, the possibility of reducing $AB$ to $D_j$ is equal to $\frac{N_j}{\sum_{1\leq i \leq k} N_i}$. Therefore, it needs runtime calculation to decide which rule to take.   
 
In the same way, instances of (\ref{eq_prec_without_p}) can be divided in two groups
\begin{equation}
(A, B, C, \prec, slf_{A, B, C}, N_s) \quad or \quad (A, B, C, \succ, rlf_{A, B, C}, N_r)
\end{equation}
It means at the context between $AB$ and $C$, the computer has noted $N_s$ times of shift and $N_r$ times of reduce done by the human parser. This makes the generated grammar $G$ a probabilistic one. When the (computer) parser of $G$ meets the same context in any future application, it may perform a shift (reduce) with probability $\frac{N_s}{N_s+N_r}$ ($\frac{N_r}{N_s+N_r}$). The decision will be made during compile time.

\section{Iterative Grammar Model development: Learn the Second Representation}
\label{sec_secondrep}

In \autoref{sec_textembedding}, we have shown how to embed all texts of $D$ in a $4 \times w$ dimensional space, where each lexical core is mapped to a vector called core and each text is mapped to an impact which is the sum of all its core vectors. In this section, we will show how to learn the target representation--an attributed rhetorical structure grammar. This will be illustrated in \autoref{alg_2_learn_arsg}.

\begin{algorithm}[h] 
\caption{Learn an attributed rhetorical structure grammar} 
\label{alg_2_learn_arsg}
\begin{algorithmic}[1] 
\Require 
a set $DS$ of natural language texts from an application domain $D$.
\State Perform \autoref{alg_1_embedtexts} to embed all texts of $DS$ and their lexical cores to two sets of vectors (impacts and cores) in a $4 \times w$ dimensional space $R$;
\State Perform Algorithm 3 to select a most representative subset $S1$ of $S$, where $S$ is the set of all impacts, for training and learn an initial model (a starting attributed rhetorical structure grammar);
\State Perform \autoref{alg_4_learn_and_improve} (recursively) to attach labels to $S2 = S-S1$ and to learn the target model (an attributed rhetorical structure grammar) for the whole set $S$ (thus also for whole $DS$).
\end{algorithmic} 
\end{algorithm}

According to different strategies of representative selection, Algorithm 3 contains two alternative versions: \autoref{alg_31_learn_model_control} and \autoref{alg_32_learn_model_percentage}.

Besides \autoref{alg_31_learn_model_control} for learning an initial model with a fixed threshold $q$ of representative selection, we provide also its alternative version \autoref{alg_32_learn_model_percentage} which selects representatives based on number of population in each cube. This tactic is more suitable for the case where the size of cubes is rather big and the impacts within most cubes are rather far from each other.  In \autoref{alg_32_learn_model_percentage}, the steps are the same as \autoref{alg_31_learn_model_control} except its Step \ref{line_q_in_alg_31}, which is replaced by the new Step \ref{line_q_in_alg_32}. 

\begin{algorithm}[h] 
\setcounter{algorithm}{30}
\caption{Learn an initial model using a control number as threshold} 
\label{alg_31_learn_model_control}
\begin{algorithmic}[1] 
\Require 
a set $S$ of impacts distributed in the $4 \times w$ dimensional space $R$.
\State Determine a $4 \times w$ dimensional cube $C$ with smallest edge lengths $C_1, C_2, C_3$ and $C_4$, such that $C$ is just enough large to contain the whole set $S$;
\State  Determine four positive integers $I_1, I_2, I_3$ and $I_4$ to divide each cube in smaller cubes such that the lengths of small cube edges are $C_1/I_1,C_2/I_2,C_3/I_3$ and $C_4/I_4$ resp.;
\State Count the numbers of impacts in each small cube;
\State Calculate the divergence $Div(imp) =\sum distance(cube(a), cube(b))$ of each impact $imp$, where the sum covers Euclidian distances of all pairs $(a, b)$ of cores composing imp with $a\neq b$. $cube(x)$ is the coordinates of the cube containing $x$;
\State Determine a positive integer $q$ such that from each cube randomly $q$ impacts are selected (or $p$ impacts if it contains only $p < q$ impacts), where the probability of selecting an impact is its divergence divided by the sum of divergences of all impacts in the same cube. Remove them from the cubes;
\label{line_q_in_alg_31}
\State Let $S1(S2)$ be the set of all these (other) impacts, $DS1(DS2)$ be those domain texts embedded to $S1(S2)$. $DS1$ is now the most representative set of texts and $S1$ the most representative impacts of $S$;
\State Use $DS1$ as training set for performing Procedure 1-2 in \cite{lu2019attributed} to build an initial model (a starting attributed rhetorical structure grammar).
\end{algorithmic} 
\end{algorithm}

\begin{algorithm}[h] 
\caption{Learn an initial model using a percentage value as threshold} 
\label{alg_32_learn_model_percentage}
\begin{algorithmic}[1] 
\State Determine a positive integer $q$ such that from each cube randomly $\left\lceil {\frac{p}{q}} \right\rceil$ impacts are selected where $p$ is the total number of impacts in this cube, where the probability of selecting an impact is its divergence divided by the sum of divergences of all impacts in the same cube. Remove them from the cubes;
\label{line_q_in_alg_32}
\end{algorithmic} 
\end{algorithm}

Note that there are yet other strategies (beyond \autoref{alg_31_learn_model_control} and \autoref{alg_32_learn_model_percentage}) for selecting the most representative set of impacts for the initial model. It depends on the concrete situation and programmer's decision.

The basic idea of this step is to cluster all unlabeled (unparsed) texts to the labeled ones according to their mutual distances. The closeness of two texts is measured according to the difference of their impacts in $R$. In some areas of machine learning research, it is not always necessary or possible to train the whole immense data corpus relevant to the current topic. We assume that there is a limit controlling the learning algorithm. This limit may be a concrete number (or a percentage value) of how much data at most should be trained from the whole data corpus. This is important because as we said above training a `label' in our case may be quite expensive. We will discuss practical cases below in \autoref{sec_experiments}. In particular, only the training of those impacts mentioned in Step \ref{line_q_in_alg_31} of \autoref{alg_31_learn_model_control}, Step \ref{line_q_in_alg_32} in \autoref{alg_32_learn_model_percentage} and Step \ref{line_select_k_in_alg4} in \autoref{alg_4_learn_and_improve} should be counted for this threshold because they need manual processing of human.

\begin{algorithm}[h] 
\setcounter{algorithm}{3}
\caption{Incrementally learn and improve the model} 
\label{alg_4_learn_and_improve}
\begin{algorithmic}[1] 
\Require 
Two constants $\varepsilon>0$ and $k>0$.
\State Define distance between any impact set $M$ and impact $e$ as: $|M, e| = inf |e-e'|$ for all $e'\in M$;
\State Let $S1$ be the set of impacts selected in \autoref{alg_31_learn_model_control} or \autoref{alg_32_learn_model_percentage}, $S2$ be the set of remaining impacts;
\While {the training limit is not yet reached}
\State Let $S12 = \{(e', e) | e'\in S2, \exists e \in S1, |e – e'| <\varepsilon \}$;
\State If $S12=\emptyset$ then \textbf{goto} Step \ref{line_step_10};
\State Call \autoref{alg_5_enhance_model} ($S12$) to build an enhanced model;
\State Let $S2' = \{e'|\exists (e', e)\in S12\}$;
\State $S1=S1\bigcup S2', S2=S2-S2'$;
\State Let $k' = min \{|S12|, k\}$;
\label{line_step_10}
\If {$k'>0$}
\State Select $S3\subseteq S2$, such that $|S3| = k'$, $\forall e \in S3,e' \in S2-S3, |S1,e|\ge |S1,e'|$;
\label{line_select_k_in_alg4}
\State Call \autoref{alg_6_increase_model} ($S3$) to build an increased model;
\State $S1=S1\bigcup S3, S2=S2-S3$.
\EndIf
\EndWhile
\end{algorithmic} 
\end{algorithm}

\begin{figure}[t]
\centering
  \subfigure[First round]{\includegraphics[width=2.3in]{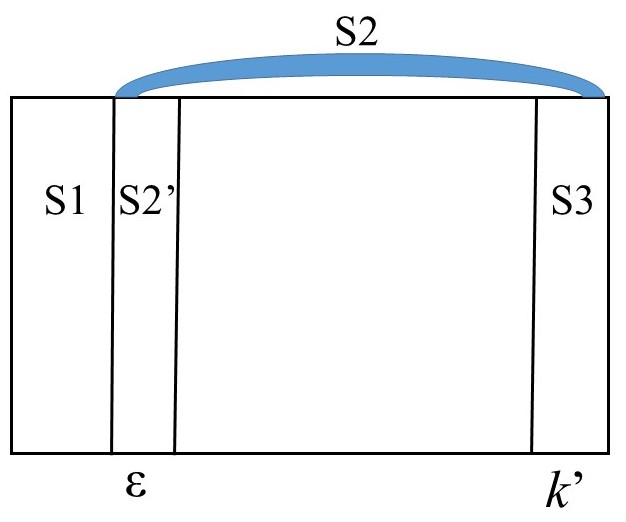}}
  \subfigure[Next round]{\includegraphics[width=2.5in]{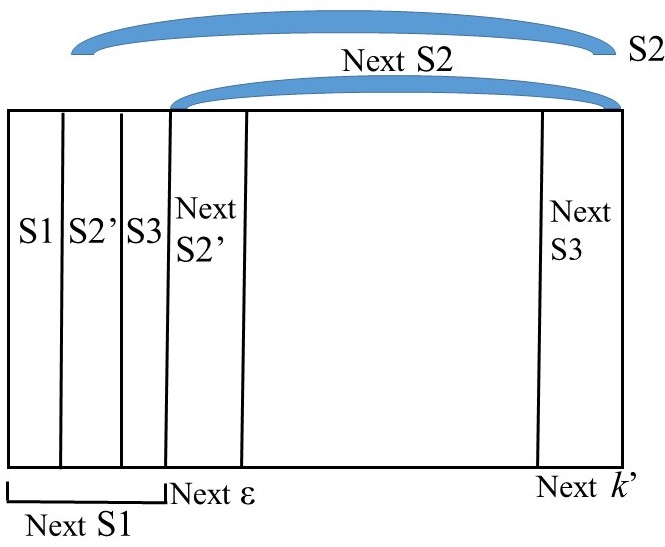}}
\caption{Learn the grammar iteratively-the space division}
\label{fig_4_learning_iter}
\end{figure}

The spatial division of impacts during \autoref{alg_4_learn_and_improve}'s running is illustrated in \autoref{fig_4_learning_iter}.

In \autoref{alg_5_enhance_model}, we denote that text which has produced impact $e$ as $text(e)$.

\begin{algorithm}[h] 
\caption{Enhance the model} 
\label{alg_5_enhance_model}
\begin{algorithmic}[1] 
\Require $S12$.
\For {each $(e',e)\in S12$}
\For {each grammar rule $r$ of $text(e)$}
\State number of instances $(r)++$;
\EndFor
\EndFor
\end{algorithmic} 
\end{algorithm}

\begin{algorithm}[h] 
\caption{Increase the model} 
\label{alg_6_increase_model}
\begin{algorithmic}[1] 
\Require $S3$.
\State Perform the actions described in \autoref{sec_getinitalgrammar} with $DS1 = \{text (e) | e \in S3\}$ to get a parsing tree and the corresponding set of grammar rules.
\end{algorithmic} 
\end{algorithm}

\begin{figure}[t]
\centering
\includegraphics[width=0.7\textwidth]{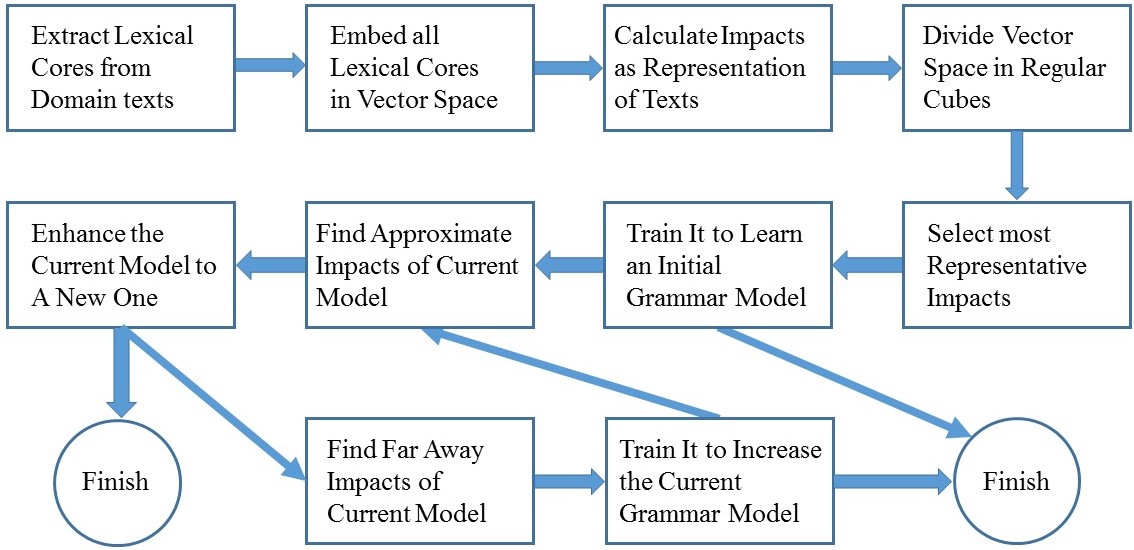}
\caption{Semi-supervised Behavioral Representation Learning: An Overview}
\label{fig_5_ssbrl_overv}
\end{figure}

The functions of these algorithms can be illustrated in \autoref{fig_5_ssbrl_overv}. 

\begin{figure}[t]
\centering
\includegraphics[width=0.7\textwidth]{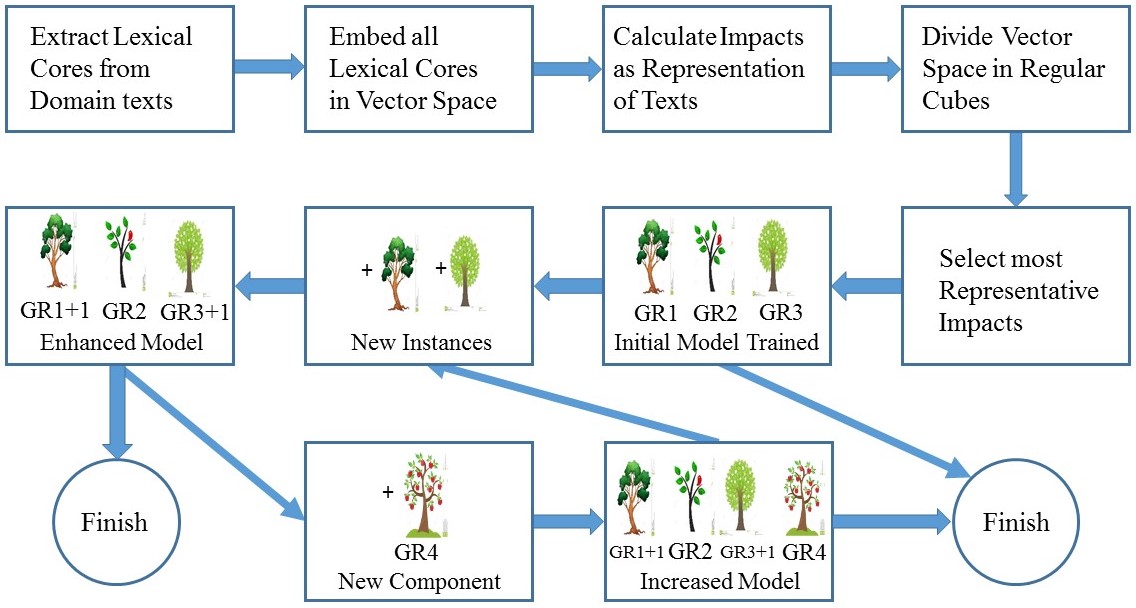}
\caption{Learn the Grammar Iteratively: The second representation (Compare it with \autoref{fig_5_ssbrl_overv})}
\label{fig_6_ssbrl_detail}
\end{figure}

\autoref{fig_6_ssbrl_detail} is a more illustrative form of \autoref{fig_5_ssbrl_overv}, where $GR$ means a set of grammar rules, $GR1+1$ means the number of instances of each rule in set $GR1$ will be increased by 1. Notice please the difference between model enhancement and model increase in \autoref{fig_5_ssbrl_overv}. In model enhancement, the numbers of different grammar rule types are unchanged. Only the numbers of rule instances are subject to change (increase). However, in case of model increase there will be possibly new rule types generated, as we can see in \autoref{fig_6_ssbrl_detail}.

Note that the concept of impacts is different from the concept centroid used in other literature such as Radev et al. \cite{radev2004centroid}, where centroid is defined as a group of words that statistically represent a cluster of texts, while impacts in this paper are defined as vector sums of cores in the embedding space.

\section{Multiple Domain Texts Summarization from Rhetorical Trees}
\label{sec_mdts}

In this section, we show how the two representations discussed above work together to help automatic domain texts summarization. First we mention an algorithm in a previous paper of us for making summary of a single text \cite{lu2019attributed}, where we established a grammar model for representing a text with its parsing tree which is an attributed rhetorical structure tree.

\begin{algorithm}[h] 
\caption{Summarize a Single Domain Text (a sketch)} 
\label{alg_7_sum_single}
\begin{algorithmic}[1] 
\Require A rhetorical structure tree and the number $n$ of requested summary length.
\State Traverse the tree according to the following principles;
\State It starts from the root of the tree;
\State It is top-down. Once a node is traversed, the next one to be traversed is its child node whenever such node exists;
\State It is nucleus first. When the traversal is going down, the nucleus child will be selected first;
\State It keeps going down until reaching a leaf which is the next significant node which will be outputted;
\State It is balanced. Whenever reaching a leaf, it restarts traversal from the highest brother node of its ancestor.
\State It stops when the requested number of EDUs have been outputted.
\end{algorithmic} 
\end{algorithm}

\begin{figure}[t]
\centering
\includegraphics[width=0.5\textwidth]{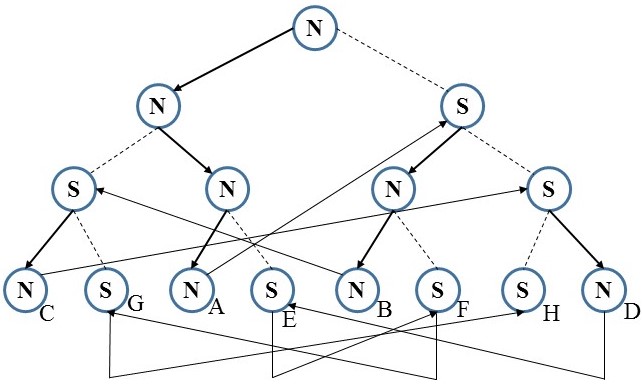}
\caption{Balanced nucleus-first search on the rhetorical tree}
\label{fig_7_sum_single}
\end{figure}

\autoref{alg_7_sum_single} performs a nucleus-first search on the rhetorical tree to obtain a priority sequence of EDUs (leaves of the tree). The number beside each leaf shows its priority alphabetically in summary sequence selection. See \autoref{fig_7_sum_single}.

In the above sections we have described a novel technique of how to learn a natural language grammar in a semi-supervised way for multiple text summarization. In the following we will show that even after the grammar is learned, there may be also tricks of applying the grammar during real summarization scripts.

First, we acknowledge that when the quantity of texts to be summarized is immense or when the scopes of the texts are very distributed, we may want to grasp and summarize only those texts which provide the mainstream information in the given domain. This leads to the strategy of given up the isolated pieces of information and concentrating on the main stream information contained in a subset of given texts to be summarized. For this purpose, we again make use of the embedding technique to select the most representative texts, as shown by the following \autoref{alg_8_sum_mds}.

\begin{algorithm}[h] 
\caption{Summarize Multiple Domain Texts} 
\label{alg_8_sum_mds}
\begin{algorithmic}[1] 
\Require The ARSG $G$ learned in \autoref{alg_2_learn_arsg}, a set $T$ of texts to be summarized from the same domain $D$, and a requested length $n$ (EDUs) of extractive content summary of $T$.
\State Call \autoref{alg_1_embedtexts} to embed all lexical cores of $T$ to the core space (space of embedded lexical cores) and impact space (space of impacts calculated from the cores); 
\State Call \autoref{alg_31_learn_model_control} (with the modification that in each cube $q$ always equals to 1 and each selected impact is attached with the number of impacts in that cube as weight of it) to determine a (most representative) ordered sequence RES of impacts from the impact space, where the order is determined by the weights of impacts;
\label{line_step_5_in_alg8}
\State Remove all impacts not belonging to $RES$ from the impact space;
\State Use $G$ to parse the set $TES$ of all texts which were embedded to $RES$ to get a priority sequence $STr$ of attributed rhetorical trees, where the priority of $text(e)$ is determined by the weight of impact $e$ in Step \ref{line_step_5_in_alg8}. 
\State Call \autoref{alg_7_sum_single} to generate a priority sequence $SEd(i)$ of EDUs for each text $t_i$ of $T$ with the help of the rhetorical trees $Tr$;
\State Use a priority function $f(x_i, y_j)$ to calculate the synthesized priority of $j$-th EDU $d_j$ of $i$-th text $t_i$ in $TES$, where $x_i$ is the $STr$ priority and $y_j$ the $SEd(i)$ priority.
\State The summarization EDUs will be selected according to and in the order of the values of priority function $f$.
\end{algorithmic} 
\end{algorithm}

It is worth noting that different from the usual approaches of summarizing all texts of a given text set, our approach evaluates the significance of each given text and summarizes only those texts above some significance degree. By significance of a text we mean its representativeness. This strategy makes the approach more efficient, in particular in case where the amount of texts to be summarized is immense. The number $n$ in \autoref{alg_8_sum_mds} is the number of EDUs rather than the number of words, which is different from traditional text summarization approaches since our method is based on lexical cores and RST techniques.

Note that it is possible to process the usual problems which may appear in the summary, such as information redundancy and/or inconsistency, based on our technique. But since the technique of multiple text summarization does not stay in the focus of our attention in this paper, we will consider them in later publications.

\section{Experiments, Evaluation and Results}
\label{sec_experiments}

It is very expensive to create a large parallel summarization corpus, and the most common case is that we have many texts to summarize, but have few or no instances of summaries. To side-step this problem, in this paper, we propose SSMRBL, especially regarding behavior learning in the form of a grammatical approach towards domain-based multiple text summarization. We apply our method to publically available Chinese dataset: SogouCA\footnote{\url{https://www.sogou.com/labs/resource/ca.php}}. SogouCA is a large-scale Chinese dataset that crawled and provided by Sogou Labs from dozens of Chinese news websites, including news reports and reviews. Each text in SogouCA contains fields of ``url'', ``docno'', ``contenttitle'', and ``content''. Leveraging ``url'' information, we can categorize texts into their corresponding domain.

In this work, we choose the domain $D$ as ``Finance''. The texts of ``Finance'' domain were selected as our experimental dataset in this work. Texts in this domain are almost news reports and comments. We did preprocessing including delete empty or very short lines, ignore extreme long lines, etc. Unlike English, to manipulate text at the word level, word segmentation is needed for Chinese text processing. We used HanLP \cite{hanlp} for Chinese word segmentation, part-of-speech (POS) tagging and named entity recognition (NER), which is a Chinese natural language processing tool. After preprocessing, the statistics of texts are listed in \autoref{tb_data_stat}.

\begin{table}
\caption{The statistics of texts of ``Finance'' domain in SogouCA}
   \centering
   \begin{tabular}{p{0.11\textwidth}p{0.19\textwidth}p{0.19\textwidth}p{0.2\textwidth}p{0.15\textwidth}}
   \toprule
   The number of texts & The number of components a (agent) & The number of components o (object) & The number of components s (state change) & Average number of EDUs per text \\
   \midrule
    261,137 & 2,865 & 5,469 & 3,308 & 21.63 \\
   \bottomrule
\end{tabular}
\label{tb_data_stat}
\end{table}

We first extract lexical cores for each text in an automatic way. Named entity recognition techniques and regular expressions were used as tool to detect time. The other three components (i.e. agent, object, and state change) were recognized by the POS tag and keyword extraction algorithms. Then, they were used to scan each text and segment it into an EDU sequence. The average number of EDUs after segmentation are also shown in \autoref{tb_data_stat}. In \autoref{tb_data_stat}, ``the number of components'' of each type denotes the number of unique components that appear in SogouCA texts. The total number of unique recognized times in all texts is 7,605, which is larger than other three components. These four components were employed to train the embedding that embed all lexical cores to vector space.

We split the dataset into training, validation, test sets according to the proportion of 8:1:1, resulting in the number of texts in each sets are 208,910, 26,114 and 26,113 respectively. We trained our model on sets of instances created from the training set, tuned hyper-parameters using instances from the validation set and tested our model on test set.

We then learn the first representation: text embedding, according to \autoref{alg_1_embedtexts}. Our model was trained with Adam optimizer \cite{kingma2014adam} with the initial learning rate as 0.001. In the margin-based objective function of (\ref{eq_emb_loss}), the margin $\gamma$ was set to 2.0 and the dissimilarity measure $d$ was set to the $L_1$-norm. The dimension of time, agent, object, and state change embeddings were all set to $w=150$. Thus the core space is a 600 dimensional space. Finally, the impact of each text was calculated as the sum of all its lexical core embeddings, which is considered as the embedded mapping of the text.

After the above procedure, each lexical core was embedded into the core space and each text was represented by the impact in the impact space. For multi-text summarization, what is missing is to determine topic-similar clusters, each of which will contribute to the multiple text source for generating a summary. Intuitively, two texts are topic-similar if their corresponding impacts are near to each other in the impact space with respect to Euclidian distance. We have validated this using a keyword-based method: (1) We randomly sampled a bigger cube that contains 997 impacts of different texts; (2) We applied RAKE (Rapid Automatic Keyword Extraction) \cite{rose2010automatic}, an unsupervised and domain-independent method for extracting keywords from individual texts, to extract keywords from each text; (3) The statistical results show that the closer two impacts are, the more common keywords their corresponding texts share. The average text similarity recall of any two closer impacts achieved 64.6\%. We have quantitatively proved our assumption. In the following summarization step, based on the impacts, we will classify texts into small clusters, each of which has approximate 120 texts.

According to \autoref{alg_2_learn_arsg}, we next learn the target representation (i.e. an ARSG). We have assumed in \autoref{sec_secondrep} that there is a threshold controlling how much data should be trained from the whole dataset in the learning procedure, which may be a concrete number (\autoref{alg_31_learn_model_control}) or a percentage value (\autoref{alg_32_learn_model_percentage}). To validate this assumption, we tested different combination of parameters. We set $q=3$, $k$ among $\{25, 50\}$. When doing ``Picking near-Training far'' iteration according to \autoref{alg_4_learn_and_improve}, the quantitative results (the number of new labeled texts) are shown in \autoref{tb_3_quan_results}. Note that ``Picking near $K$'' denotes the number of new labeled texts in $K$-th iteration, the same as ``Training far $K$''. To show our experiment more intuitive, we list the results that $K$ among $\{1, 3, 6\}$. From this table, we can see that more labeled texts can be derived when using more labeled training data to build the initial grammar model. Moreover, in the ``Training far'' steps, $k=50$ can still obtain considerable number of labeled texts when both using the concrete number or the percentage value as the threshold. In the following experiments, we will set $k=50$.

\begin{table}
\caption{Quantitative results when doing ``Picking near-Training far'' iteration}
   \centering
   \begin{tabular}{p{0.15\textwidth}p{0.15\textwidth}p{0.15\textwidth}p{0.15\textwidth}p{0.15\textwidth}}
   \toprule
   & \multicolumn{2}{c}{Concrete threshold (\autoref{alg_31_learn_model_control})} &\multicolumn{2}{c}{Percentage threshold (\autoref{alg_32_learn_model_percentage})} \\
   Parameters & $q=3, k=25$ & $q=3, k=50$ & $q=3, k=25$ & $q=3, k=50$ \\
   \midrule
    Initial & 97 & 97 & 166 & 166 \\
    Picking near 1 & 136 & 136 & 217 & 217 \\
    Training far 1 & 23 & 44 & 23 & 45 \\
    Picking near 3 & 1,039 & 1,039 & 2,266 & 2,266 \\
    Training far 3 & 23 & 41 & 22 & 46 \\
    Picking near 6 & 6,297 & 6,297 & 8,547 & 8,547 \\
    Training far 6 & 25 & 49 & 24 & 49 \\
    The number of all labeled texts & 180,971 & 172,975 & 189,264 & 186,637 \\
    The scale of ultimate grammars & $PPR(AT)$: 9,145 $PF(AT)$: 9,812 & $PPR(AT)$: 9,068 $PF(AT)$: 9,617 & $PPR(AT)$: 9,236 $PF(AT)$: 10,318 & $PPR(AT)$: 9,205 $PF(AT)$: 9,878\\
   \bottomrule
\end{tabular}
\label{tb_3_quan_results}
\end{table}

The next step is to apply \autoref{alg_8_sum_mds} for multi-text summarization. When measuring the quality of generated summary, the commonly used evaluation metric for text summarization is ROUGE \cite{lin2004rouge}. ROUGE evaluates $n$-gram co-occurrences between summary pairs. It works by comparing an automatically produced summary against a set of reference summaries. Since our method is semi-supervised, we do not only assume there is only a small number of RST-labeled texts, but assume generating summary from multiple texts about the same topic without any parallel texts-summary pair. We defined proxy metrics to evaluate our generated summaries without example summaries. To build a quantization standard, we consider the following automatic statistics
\begin{enumerate}
\item \textbf{Sentiment polarity accuracy.} A summary should reflect and be consistent with the overall sentiment of the original texts. To address the problem that resources for Chinese sentiment analysis are limited, we translated the title of each text into English by open-source machine translation services, and then identified the sentiment polarity of the English version by directly leveraging English-oriented algorithm, whose result was taken as the sentiment of Chinese text. We then trained a CNN-based sentiment analyzer that given a text, predicts the sentiment polarity among $\{-1,0,1\}$. For each summary, we compute the accuracy of whether the sentiment analyzer's predicted polarity is equal to the average polarity of all the texts in the cluster.
\item \textbf{Novelty of summary EDUs.} The similarities between each two of the summary EDUs should be lower in case of redundancy. We compute the redundancy by averaging the F1 scores of ROUGE-1, ROUGE-2 and ROUGE-L between each two of the EDUs, and then average these averaged scores. The novelty of summary EDUs is computed as:
\begin{equation*}
Novelty = 1 - \mathop {avg}\limits_{a,b \in EDUs,a \ne b} ({R_1}(a,b) + {R_2}(a,b) + {R_L}(a,b))/3
\end{equation*}
\item \textbf{The appearance of named entities.} The summary should contain more named entities (NEs) that the texts in the cluster have. We obtain this score by computing the number of unique NEs appeared in the original texts divided by the number of unique NEs appeared in the summary.
\item \textbf{Word overlap score.} Word overlap score can be used to measure how much the generated summary encapsulates the original texts. We compute this score by using the F1 scores of ROUGE-1 between the summary and each text in the cluster and then average these scores.
\end{enumerate}

Formally, the score of a generated summary of multiple texts is computed as
\begin{equation}
Score = \alpha S + \beta N + \gamma E + \omega W
\end{equation}
where $S, N, E, W$ represents sentiment polarity accuracy, novelty of summary EDUs, the appearance of named entities, and word overlap score respectively; $\alpha, \beta, \gamma, \omega$ are weights, which were empirically set to $\{20, 25, 25, 30\}$, thus the range of score is $[0, 100]$.

Since the initial labeled texts are obtained manually, which is complex and time-consuming. We also conducted experiments on different number of labeled texts for initial grammar learning, and on different total number of training texts. We set the size of S1 among $\{60, 90, 120, 150, 180\}$, and set the number of training texts among $\{10,000, 50,000, 100,000, All\}$. For each combination of parameters, we applied \autoref{alg_2_learn_arsg} to learn an ARSG and applied \autoref{alg_8_sum_mds} to the clustered texts. We set the parameter $n=10$ in \autoref{alg_8_sum_mds}. \autoref{tb_sum_n10} shows the scores of generated summaries which reflect the effect of initial labeled texts on the performance of summary.

\begin{table}
\caption{The effect of initial labeled texts on the performance of summary ($n=10$)}
   \centering
   \begin{tabular}{p{0.08\textwidth}p{0.07\textwidth}p{0.07\textwidth}p{0.07\textwidth}p{0.07\textwidth}}
   \toprule
   $|S1|$ & 90 & 120 & 150 & 180 \\
   \midrule
10,000 & 46.2 & 69.7 & 71.2 & 71.6  \\
50,000 & 48.5 & 72.3 & 74.3 & 75.2 \\
100,000 & 52.3 & 77.6 & 78.4 & 80.5 \\
All & 52.7 & 78.4 & 80.1 & 79.7 \\
   \bottomrule
\end{tabular}
\label{tb_sum_n10}
\end{table}

When using 100,000 training texts and 120 labeled texts, our model can obtain relatively good performance, since there is little promotion when adding more training texts or labeled texts, and maybe even ``misleading'' when using more training texts.

Moreover, to further assess the quality of summaries, we also conducted experiments on our test set using baselines. Since our method generates summary without any parallel texts-summary pair, we select representative unsupervised multi-text summarization approaches as the baselines, including the strong baseline Lead, graph-based method (TextRank) and deep learning based method (CNNLM+TextRank).
\begin{enumerate}
\item \textbf{Lead:} takes the first EDUs one by one in the text of the cluster until length limit, where texts are assumed to be ordered according to their weight, which has been assigned in \autoref{alg_8_sum_mds}.
\item \textbf{TextRank \cite{mihalcea2004textrank}:} builds a graph and adds each sentence as vertices, the overlap of two sentences is treated as the relation that connects sentences. Then graph-based ranking algorithm is applied until convergence. Sentences are sorted based on their final score and a greedy algorithm is employed to impose diversity penalty on each sentence and select summary sentences.
\item \textbf{CNNLM+TextRank \cite{yin2015optimizing}:} applies CNN to project sentences to continuous vector space and uses TextRank for sentence ranking. CNNLM extracts the sentence representation by CNN and combines that sentence representation with the representations of context words to predict the next word. CNNLM was trained in an unsupervised scheme, which resembles the CBOW scheme in Word2vec \cite{mikolov2013efficient}. After finishing training, a sentence adjacent graph based on cosine similarity was built for TextRank.
\end{enumerate}

We denote our method by DBMTS with 100,000 training texts and 120 labeled texts, the automated score for our model and the baselines are shown in \autoref{tab_compare_rel}.

\begin{table}
\caption{Experimental results of our model comparing with baselines}
   \centering
   \begin{tabular}{p{0.17\textwidth}p{0.1\textwidth}p{0.1\textwidth}}
   \toprule
   Model	 & $n=10$ & $n=5$ \\
   \midrule
Lead & 70.5 & 67.8 \\
TextRank & 72.3 & 69.4 \\
CNNLM+TextRank & 74.9 & 72.2 \\
DBMTS & 77.6 & 72.5 \\
   \bottomrule
\end{tabular}
\label{tab_compare_rel}
\end{table}

The DBMTS model outperforms all the baselines in our proposed evaluation metric. It also obtains a slightly promotion when $n=5$ comparing with CNNLM+TextRank, whereas the performance is better when $n=10$. From the tables we observe that our proposed method brings a good performance when small labeled texts are used with respect to our defined automated metrics.

\section{Concluding Remarks}
\label{sec_conclusion}

We note some features of our approach: Firstly, our approach is cross-representational. It is implemented in parallel both on natural language level and on embedded digital level. Secondly, since behaviorally training a natural language grammar is always a hard work, we only train a part of the text corpus to the amount of necessity. Thirdly, the training set is not arbitrary taken but optimally selected according to its representation capability which helps optimize the algorithm. Fourthly, this approach is incremental such that new data (texts) can be added at any time to generalize and improve the grammar. Fifthly, the approach is also modular since the grammar is implemented in a group of modules (rule pieces) which can be reorganized to meet different needs and situations. Sixthly, this approach is flexible since not only our ARSG, but any other grammar can be used. Seventh, this approach is even universal since its idea can be applied to any other semi-supervised learning where the training and transition of data labels is a hard work.

\section*{Acknowledgments}
This work was supported by the National Key Research and Development Program of China under grant 2016YFB1000902; and the National Natural Science Foundation of China (No. 61232015, 61472412, and 61621003).

\section*{References}

\bibliography{refs}

\end{document}